\documentclass[twoside,11pt]{article}

\usepackage{blindtext}
\usepackage{natbib}
\usepackage{jmlr2e}
\usepackage{amsmath}
\usepackage{color}

\usepackage{lastpage}

\firstpageno{1}

\begin{document}

\title{Gaussian kernel expansion with\\ basis functions uniformly bounded in  $\mathcal{L}_{\infty}$}

\author{\name Mauro Bisiacco \email bisiacco@dei.unipd.it\\
       \addr Department of Information Engineering\\
       University of Padova\\
       Padova, 35131, Italy
       \AND
       \name Gianluigi Pillonetto \email giapi@dei.unipd.it \\
       \addr Department of Information Engineering\\
       University of Padova\\
       Padova, 35131, Italy}

\editor{}

\maketitle

\begin{abstract}

Kernel expansions are a topic of considerable interest in machine learning, also because of their relation to the so-called feature maps introduced in machine learning. Properties of the associated basis functions and weights (corresponding to eigenfunctions and eigenvalues in the Mercer setting) give insight into for example the structure of the associated reproducing kernel Hilbert space, the goodness of approximation schemes, the convergence rates and generalization properties of kernel machines. Recent work in the literature has derived some of these results by assuming uniformly bounded basis functions in  $\mathcal{L}_\infty$. Motivated by this line of research, we investigate under this constraint all possible kernel expansions of the Gaussian kernel, one of the most widely used models in machine learning. Our main result is the construction on $\mathbb{R}^2$ of a Gaussian kernel expansion with weights in $\ell_p$ for any $p>1$. This result is optimal since we also prove that $p=1$ cannot be reached by the Gaussian kernel, nor by any of the other radial basis function kernels commonly used in the literature. A consequence for this kind of kernels is also the non-existence of Mercer expansions on $\mathbb{R}^2$, with respect to any finite measure, whose eigenfunctions all belong to a closed ball of $\mathcal{L}_\infty$.
\end{abstract}


\section{Introduction}


Kernels are widely used in machine learning to estimate functions from sparse and noisy data, also in virtue of the link with reproducing kernel Hilbert spaces (RKHSs) \citep{Aronszajn50,Cucker01} and Gaussian regression \citep{Rasmussen,PilAuto2007,PNAS2023}.  Their combination with Tikhonov-style regularization led to the development of powerful algorithms such as support vector machines \citep{Cortes95,Vapnik95,Drucker97} and kernel ridge regression 
\citep{PoggioMIT,Yuan2010}. Connections with control and system identification are also discussed in 
\citep{CCP:12,Carron2015,SpringerRegBook2022,Care2023}.
Popular kernels include splines \citep{Wahba:90}, the polynomial kernel \citep{Poggio75} and the class relying on radial basis functions \citep{Hastie01,Scholkopf01b}. Among the latter, a notable example is the Gaussian kernel, probably the most widely used in applications.\\
The importance of kernel machines has motivated a myriad of studies on their computational aspects and theoretical properties, e.g. \citep{Blanchard2018,Ref12,MathFoundStable2020,StableTAC2024,Kostic2022,Wang2022,Campi2021}. Among them kernel expansions are a topic of considerable interest, also in view of their relationships with the so called feature maps introduced in learning theory \citep{Scholk99,Minh:09}. Characteristics of basis functions and weights, defined by kernel eigenfunctions and eigenvalues in the Mercer setting \citep{Sun05,Takh2023}, give insights into the goodness of approximation schemes, the convergence rates and generalization properties of regularized algorithms, the structure of the associated RKHS. Regarding this last issue, an important contribution is  \citep{IS2006} where an explicit characterization of the RKHS induced by the Gaussian kernel was provided.\\ 
\indent Some works in the literature have studied regularized estimators assuming that the kernel admits an expansion through basis functions uniformly bounded in $\mathcal{L}_\infty$ \citep{Fischer2020,Laff05,Mendelson2010,Pillaud2018,Steinwart2009}. The difficulty of obtaining this kind of feature maps by the Mercer theorem is well discussed in \citep{Zhou:02} and \citep{Minh:09}[Section 3]. In \citep{Giaretta23} it was also shown that the existence of any expansion, not necessarily Mercer, enjoying such property leads to new consistency results for rather general (non-stationary) data generators if the weights suitably decay to zero. For instance, 
convergence (in probability) in the space of continuous functions is ensured if the expansion coefficients lie in $\ell_1$. 
Motivated by this line of research, this paper then investigates the existence of such feature maps with a special focus on the Gaussian kernel.\\ 
Our analysis depends on the nature of the domain $X$ where the kernel is defined. Over any bounded domain we easily prove the existence of the Gaussian kernel expansion with basis functions uniformly bounded in $\mathcal{L}_\infty$ and weights in $\ell_1$. Over $\mathbb{R}^2$ we instead show that this property can never hold for any 
radial basis functions kernels commonly used in the literature, including e.g. the Gaussian, Laplace and Cauchy kernel. 
For this kind of kernels, this result also implies the non-existence of Mercer expansions on $\mathbb{R}^2$, with respect to any finite measure, whose eigenfunctions all belong to a closed ball of $\mathcal{L}_\infty$.
The analysis is then specialized to the Gaussian kernel case by constructing an expansion 
with weights in $\ell_p$ for any $p>1$ which is therefore optimal in the light of the above result. The next two sections formally report these results and their proofs.

\section{The main result}

We use $K$ to denote a positive semidefinite kernel. It is a symmetric function over $X \subseteq {\mathbb R}^2$ such that, given any integer $n$, scalars $\{a_i\}_{i=1}^n$ and vectors $\{x_i\}_{i=1}^n$ in $X$, one has
$$
\sum_{i=1}^n \sum_{j=1}^n a_ia_j K(x_i,x_j) \geq 0. 
$$
The following definition introduces the expansions of interest within this article where here, and in what follows, convergence is always intended in the pointwise sense.

\begin{definition}
The kernel $K$ admits a $(p,\infty)$-expansion if 
\begin{equation}\label{DefExpansion}
K(x,y)=\sum_{k=1}^{+\infty} \ \lambda_k\psi_k(x)\psi_k(y), \quad \lambda_k \geq 0
\end{equation}
where the sequence of expansion coefficients $\lambda_k$ 
belongs to $\ell_p$ and all the $\psi_k$ are linearly independent functions which belong to a ball of $\mathcal{L}_{\infty}$, i.e. there exists $M$ independent of $k$
such that 
$$
\| \psi_k \|_{\infty} \leq M < +\infty \quad \forall k.
$$
\end{definition}

\noindent Our main result is then stated below.

\begin{proposition}\label{Main}
Let $X$ be any bounded subset of $\mathbb{R}^2$.   
Then, the Gaussian kernel admits a $(1,\infty)$-expansion.
Let instead $X=\mathbb{R}^2$. Then, any non null radial basis functions kernel $K(x,y)=F(|x-y|)$, with $F$ infinitesimal as $|x-y|$ grows to infinity,  does not admit a $(1,\infty)$-expansion. The Gaussian kernel however admits a $(p,\infty)$-expansion for any $p>1$.
\end{proposition}

\noindent Under conditions discussed e.g. in \citep{Sun05}, Mercer theorem
provides an expansion \eqref{DefExpansion} over $\mathbb{R}^2$ under the measure
$\mu$ with
$$
\lambda_k \psi_k(x) = \int K(x,y)\psi_k(y)d\mu(y)
$$
and the $\psi_k$ orthonormal in $\mathcal{L}^2_{\mu}$. 
If $K$ is a radial basis functions kernel and the measure $\mu$ is finite, for example a probability measure such as
in statistical learning \citep{Smale2007}, the sequence $\lambda_k$ belongs to $\ell_1$. For instance, using the Gaussian measure, the eigenfunctions of the Gaussian kernel are Hermite functions \citep{Rasmussen}[Section 4.3.1]. 
The following corollary of Proposition \ref{Main} is then immediately obtained.

\begin{corollary}
 
Let $X=\mathbb{R}^2$ and $\mu$ be a finite measure.
Then, there does not exist any Mercer expansion
of non null radial basis functions kernel $K(x,y)=F(|x-y|)$, with $F$ infinitesimal as $|x-y|$ grows to infinity, whose eigenfunctions all belong to a ball in $\mathcal{L}_{\infty}$.
\end{corollary}

\section{Proof of the main result}

Our starting point is the following expansion of 
the Gaussian kernel of width $\eta$ taken from 
\citep[Eq. 15]{Minh:09} which holds over any subset $X$ of $\mathbb{R}^2$:
\begin{equation}\label{BasicExp}
\exp\Big(-\frac{(x-t)^2}{\eta}\Big) = \exp\Big(-\frac{x^2}{\eta}\Big)\exp\Big(-\frac{t^2}{\eta}\Big) \sum_{k=0}^{+\infty} \frac{(2/\eta)^k}{k!} x^k t^k.
\end{equation}

\subsection{The case of bounded domain $X$}

Without loss of generality let $X={\mathcal Q}_N$, where
${\mathcal Q}_N$ is the square of sides $[0,N]$ which contains any bounded set $X\subset {\mathbb R}^2$ for $N$ large enough. Let also $\eta=1$, otherwise we can replace $K(x,t)$ by $K(ax,at)$, with $a>0$, without modifying the properties obtained below. It holds that
$$
K(x,t)=\sum_{k=0}^{+\infty} \ \psi_k(x)\psi_k(t)
$$
with
$$
\psi_k(x):=\sqrt{\frac{2^k}{k!}}x^ke^{-x^2}.
$$
The non-negative function $\psi_k(x)$ has a local maximum at $x=\sqrt{\frac{k}{2}}$ larger than $N$ if $k$ is large enough. 
So for large $k$ its maximum value is attained at the boundary $x=N$ and turns out to be
$$
\sqrt{\frac{2^k}{k!}}N^ke^{-N^2}.
$$
Let us now consider $h_k(x):=k\psi_k(x)$, so that
\begin{eqnarray*}
    K(x,t)&=&\sum_{k=0}^{+\infty} \ \frac{1}{k^2} h_k(x)h_k(t)\\
    &:=&\sum_{k=0}^{+\infty} \ \lambda_k h_k(x)h_k(t).
\end{eqnarray*}

\noindent Clearly, one has $\sum_{k=0}^{+\infty} \lambda_k<+\infty$ 
and the maximum of $h_k(x)$, for large $k$, is 
$$
\|h_k\|_{\infty} = k\sqrt{\frac{2^k}{k!}}N^ke^{-N^2}.
$$
It suffices to show that
$$
\lim_{k\rightarrow+\infty} \ k\sqrt{\frac{2^k}{k!}}N^ke^{-N^2}=0
$$
to prove that $h_k(x)$'s are uniformly bounded. By Stirling's formula \citep{Bender2009},
still for $k$ large enough one has
\begin{eqnarray*}
k\sqrt{\frac{2^k}{k!}}N^ke^{-N^2} &\simeq& \frac{e^{-N^2}k}{\sqrt[4]{2\pi k}}\left(\frac{2eN^2}{k}\right)^{\frac{k}{2}} \\
&=&\frac{e^{-N^2}k}{\sqrt[4]{2\pi k}}\ e^{-k\left(\frac{ln(k)}{2}-\frac{ln(2eN^2)}{2}\right)}\\
&\le& \frac{e^{-N^2}k}{\sqrt[4]{2\pi k}}\ e^{-k\left(\frac{ln(k_0)}{2}-\frac{ln(2eN^2)}{2}\right)} \quad \forall k \ge k_0.
\end{eqnarray*}
Then, we can find $k_0>2eN^2$ ensuring the existence of $A,B>0$ leading to the upper bound 
$$
\|h_k\|_{\infty}  \le Ak^{\frac{3}{4}}e^{-Bk}, \ \ k \ge k_0
$$
which tends to zero as $k$ grows to infinity.

\subsection{The case $X=\mathbb{R}^2$: non-existence of $(1,\infty)$-expansions for all the commonly used radial basis functions kernels}

Consider now the class of radial basis functions (RBF) kernels over ${\mathbb R}^2$
$$
K(x,y):=F(|x-y|)
$$
satisfying
$$
F(0)=1, \ \ \lim_{x\rightarrow\infty} \ F(x)=0.
$$
As said, the limit condition is satisfied by any RBF kernel used in the applications. The assumption $F(0)=1$ is instead made w.l.o.g. and  excludes only the null kernel $K(x,y)=0$.\\

\noindent Given $\epsilon>0$, there exists a sequence $\{a_n\}_{n\ge 1}$ such that
$$
|F(x)|<\frac{\epsilon}{2^n} \ \mbox{for any} \ x\ge a_n
$$
so that
$$
\sum_{i=1}^{+\infty} \ |F(x_i)|<\epsilon\left(\frac{1}{2}+\frac{1}{4}+\dots\right)=\epsilon \ \ \mbox{if} \ \ x_i>a_i, \ \forall i\ge 1.
$$
Consider now, associated with $K$, any function $\psi(x)$ 
that does not tend to zero. So, there exists an increasing sequence $y_n$ with the following property: for some $\delta>0$ one has
$$
|\psi(y_n)|>\delta \ \mbox{and} \ y_{n+1}-y_n>a_{n+1}.
$$
Let now $a_i=\pm\frac{1}{\sqrt{n}}$, which implies
$$
\sum_{i,j=1}^n \ a_ia_jK(y_i,y_j)=\sum_{i=1}^n \ a_i^2K(y_i,y_i)+2\sum_{i=1}^n\ \sum_{j<i} \ a_ia_jK(y_i,y_j).
$$
Regarding the first term, one has
$$
\sum_{i=1}^n \ a_i^2K(y_i,y_i)=\frac{1}{n}\sum_{i=1}^n \ F(0)=\frac{n}{n}=1.
$$
As for the second term, first notice that,
for any fixed $i$, there exist $i-1$ terms of the form $|F(y_i-y_j)|$, with $y_i-y_j>a_i$.  Therefore $|F(y_i-y_j)|<\frac{\epsilon}{2^i}$, so that $\sum_{j < i} \ |F(y_i-y_j)|<\frac{(i-1)\epsilon}{2^i}$. It follows 
that
\begin{eqnarray*}
|\ 2\sum_{i=1}^n\ \sum_{j<i} \ a_ia_jK(y_i,y_j) \ |&\le& 2\sum_{i=1}^n\ \sum_{j<i} \ |a_ia_j| \ |K(y_i,y_j)|\\
&=&\Big| \frac{2}{n}\sum_{i=1}^n \ \sum_{j < i} \ |F(y_i-y_j)| \Big|\\
&\le& \frac{2\epsilon}{n}\left(\frac{n-1}{2^n}+\frac{n-2}{2^{n-1}}+\frac{1}{4}\right)\\
&=& \frac{\epsilon}{n}\left(\frac{n-1}{2^{n-1}}+\frac{n-2}{2^{n-2}}+\frac{1}{2}\right)\\
&<&\frac{\epsilon}{n}\left(\frac{n}{2^{n-1}}+\frac{n}{2^{n-2}}+\frac{n}{2}\right)\\
&<&\epsilon.
\end{eqnarray*}
This proves that
$$
1-\epsilon<\sum_{i,j=1}^n \ a_ia_jK(y_i,y_j)<1+\epsilon
$$
and, by choosing the signs of the $a_i$'s in such a way that $\text{sign}[a_i]=\text{sign}[\psi(y_i)]$, also that
\begin{eqnarray*}
\sum_{i=1}^n \ a_i\psi(x_i)&=&\frac{1}{\sqrt{n}}\sum_{i=1}^n \ |\psi(y_i)| \\
&>& \frac{1}{\sqrt{n}}\sum_{i=1}^n \ \delta=\sqrt{n}\ \delta \\ &\Rightarrow&  \left(\sum_{i=1}^n \ a_i\psi(y_i)\right)^2>n \delta^2.
\end{eqnarray*}
Now, still considering convergence in the 
pointwise sense,
assume that, for some $\lambda_k>0$, one has
$$
K(x,y)=\sum_{k=1}^{+\infty} \ \lambda_i\psi_k(x)\psi_k(y), \ \ \lambda_k>0.
$$
Let $\delta>0$ and the $y_i$'s be chosen accordingly to what seen before for the couple $(K,\psi_k)$ (clearly, both $\delta$ and the $y_i$ depend on $k$ too). One thus has
\begin{eqnarray*}
1+\epsilon&>& \sum_{i,j=1}^n \ a_ia_jK(y_i,y_j)\\
&=&\sum_{i,j=1}^n \ a_ia_j \ \sum_{k=1}^{+\infty} \ \lambda_k\psi_k(y_i)\psi_k(y_j)\\
&=& \sum_{i,j=1}^n \ \sum_{k=1}^{+\infty} \ a_ia_j\lambda_k\psi_k(y_i)\psi_k(y_j)=\\
&=&\sum_{k=1}^{+\infty} \ \lambda_k \ \sum_{i,j=1}^n \ a_ia_j\ \psi_k(y_i)\psi_k(y_j)\\
&=& \sum_{k=1}^{+\infty} \lambda_k \ \left[\sum_{i=1}^n \ a_i\psi_k(y_i)\right]^2\\
&\ge&\lambda_k \ \left[\sum_{i=1}^n \ a_i\psi_k(y_i)\right]^2 > \lambda_k\delta^2n.
\end{eqnarray*}
One must then have
$$
n<\frac{1+\epsilon}{\lambda_k\delta^2}
$$
but this is not possible because $n$ can assume any integer value. Therefore the contradiction implies that $\psi_k(x)$ must tend to zero. This argument can be repeated for any $\psi_k$. So, summarizing, 
so far we have proven that under the stated assumptions 
$$
K(x,y):=F(|x-y|), \ F(0)=1, \ \lim_{x\rightarrow+\infty} \ F(x)=0,
$$
if the kernel admits the expansion
$$
K(x,y)=\sum_{k=1}^{+\infty} \ \lambda_k\psi_k(x)\psi_k(y), \ \lambda_k>0 
$$
then necessarily any $\psi_k(x)$ must tend to zero.\\


\noindent Assume now that the sequence $\lambda_k$ is summable and that 
$\|\psi_k\|_{\infty} \le M$. 
For any $\epsilon>0$ we can then choose $N$ such that
$$
\sum_{k=N+1}^{+\infty} \ \lambda_k<\frac{\epsilon}{2M^2} 
$$
so that
$$
\sum_{k=N+1}^{+\infty} \ \lambda_k\psi_k^2(x)<\frac{\epsilon}{2}
$$
and this implies
\begin{equation}\label{FirstEps}
\sum_{k=1}^N \ \lambda_k\psi_k^2(x) > 1-\frac{\epsilon}{2}
\end{equation}
since 
$$
1=K(x,x)=\sum_{k=1}^{+\infty} \ \lambda_k\psi_k^2(x).
$$
Since we have proved that each $\psi_k^2(x)$ must tend to zero, 
considering the finite set of functions $\psi_1,\dots,\psi_N$ there exists $x_0$ such that $x>x_0$ implies 
$$
\psi_k^2(x)<\frac{\epsilon}{2N\lambda_k}
$$
and such inequality also leads to
\begin{equation}\label{SecondEps}
\sum_{k=1}^N \ \lambda_k\psi_k^2(x) < \frac{\epsilon}{2}.
\end{equation}
The combination of \eqref{FirstEps} and \eqref{SecondEps} leads to
$$
1-\epsilon<0
$$
which is impossible because $\epsilon$ can be any positive number. 
Such contradiction thus proves that $K$ cannot admit any $(1,\infty)$-expansion and so concludes also this part of the proof.

\subsection{The best $(p,\infty)$ Gaussian kernel expansion over $\mathbb{R}^2$}

We now prove the final part of our main result 
by explicitly constructing a Gaussian kernel 
expansion which is $(p,\infty)$ for any $p>1$. 
We start providing some useful approximations.

\subsubsection{Useful approximations}

\noindent Consider
$$
f(x)=\left(\sqrt{\frac{2e}{k}}\right)^kx^ke^{-x^2}.
$$
Defining $y=x-\sqrt{\frac{k}{2}}$, one has
\begin{eqnarray*}
f(y+\sqrt{\frac{k}{2}})&\simeq&\left(\sqrt{\frac{2e}{k}}\right)^k\left(y+\sqrt{\frac{k}{2}}\right)^ke^{-\left(y+\sqrt{\frac{k}{2}}\right)^2}\\
&\simeq&\left(\sqrt{\frac{2e}{k}}\right)^k\left(\sqrt{\frac{k}{2}}\right)^k\left[\left(1+\sqrt{\frac{2}{k}}\ y\right)^{\sqrt{\frac{k}{2}}}\right]^{\sqrt{2k}}e^{-y^2}e^{-\frac{k}{2}}e^{-\sqrt{2k} \ y}\\
&\simeq&e^{\frac{k}{2}}\left[\left(1+\sqrt{\frac{2}{k}}\ y\right)^{\sqrt{\frac{k}{2}}}\right]^{\sqrt{2k}}e^{-y^2}e^{-\frac{k}{2}}e^{-\sqrt{2k} \ y}\\
&\simeq&\left[\frac{\left(1+\sqrt{\frac{2}{k}}\ y\right)^{\sqrt{\frac{k}{2}}}}{e^y}\right]^{\sqrt{2k}}e^{-y^2}\\
&:=&A.
\end{eqnarray*}
Taking the logarithm
\begin{eqnarray*}
log(A)&=&\sqrt{2k}\left[\sqrt{\frac{k}{2}}log\left(1+\sqrt{\frac{2}{k}}y\right)-y\right]-y^2\\
&\simeq&\sqrt{2k}\left[\sqrt{\frac{k}{2}}\left(\sqrt{\frac{2}{k}}y-\frac{1}{k}y^2\right)-y\right]-y^2\\
&\simeq&\sqrt{2k}\left[y-\frac{1}{\sqrt{2k}}y^2-y\right]-y^2\\
&\simeq&-2y^2
\end{eqnarray*}
from which
$$
A=e^{log(A)}\simeq e^{-2y^2}.
$$
Therefore
$$
f(x)=\left(\sqrt{\frac{2e}{k}}\right)^kx^ke^{-x^2}\simeq e^{-2\left(x-\sqrt{\frac{k}{2}}\right)^2}.
$$
Now, Stirling's formula leads to the following
useful approximation (whose accuracy improves as $k$ gets larger) for the key functions $\psi_k(x)$:
\begin{equation}
\psi_k(x)=\sqrt{\frac{2^k}{k!}}x^ke^{-x^2} \simeq \frac{f(x)}{\sqrt[4]{2\pi k}}\simeq \frac{1}{\sqrt[4]{2\pi k}}\ e^{-2\left(x-\sqrt{\frac{k}{2}}\right)^2}.
\label{Gauss}
\end{equation}

\subsubsection{The maxima sequence}

The expansion
\begin{equation}
K(x,y)=\sum_{k=0}^{+\infty} \ \psi_k(x)\psi_k(y), \ m_k:=\|\psi_k\|_{\infty}
\label{AAA}
\end{equation}
can be rewritten as
\begin{equation}
K(x,y)=\sum_{k=0}^{+\infty} \ \lambda_k \phi_k(x)\phi_k(y), \ \phi_k(x)=\frac{\psi_k(x)}{m_k}, \ \|\phi_k(x)\|_{\infty}=1, \ \lambda_k=m_k^2.
\label{BBB}
\end{equation}
Hence, (\ref{AAA}) gives rise to (\ref{BBB}) with $\phi_k$ uniformly bounded in $\mathcal{L}_{\infty}$ and with 
\begin{equation}
\sum_{k=0}^{+\infty} \ \lambda_k=\sum_{k=0}^{+\infty} \ m_k^2=\sum_{k=0}^{+\infty} \ \|\psi_k\|_{\infty}^2.
\label{CCC}
\end{equation}
Still using \eqref{BasicExp} one has
$$
K(x,y)=\sum_{k=0}^{+\infty} \ \psi_k(x)\psi_k(y), \ \psi_k(x)=\sqrt{\frac{2^k}{k!}}x^ke^{-x^2}
$$
The function $\psi_k(x)$ is either even or odd, so its analysis can be restricted to $x\ge 0$. Its maximum value is obtained at $x:=x_k=\sqrt{\frac{k}{2}}$ and, exploiting the Stirling approximation for large $k$, it holds that
$$
m_k=\sqrt{\frac{k^k}{k!}}e^{-\frac{k}{2}}\simeq\frac{1}{\sqrt[4]{2\pi k}} \ \Rightarrow \ m_k^2\simeq\frac{1}{\sqrt{2\pi k}}.
$$
Note that the sequence $\lambda_k=m_k^2$ is not summable and does not belong to $\ell_2$ either.\\
Using (\ref{Gauss}), for $0<\alpha<1$ we define $\delta_k(\alpha)$ through the following implication
$$
x_k(1\pm \delta_k(\alpha)) \ \Rightarrow \ \psi_k[x_k(1\pm\delta_k(\alpha))]=\alpha m_k, \ 0<\alpha<1
$$
which also implies
$$
x_k(1\pm\delta_k(\alpha))=x_k\pm x_k\delta_k(\alpha)\simeq x_k \pm \sqrt{-\frac{ln[\alpha]}{2}}.
$$
The last approximation shows that, for $k$ large enough, $x_k\delta_k(\alpha)$ does not depend on $k$ but only on $\alpha$. Hence, if
\begin{equation*}
y_1=x_k-\sqrt{-\frac{ln[\alpha]}{2}}, \ y_2=x_k, \ y_3=x_k+\sqrt{-\frac{ln[\alpha]}{2}} 
\end{equation*}
then
\begin{equation}
\psi_k(y_1)=\alpha m_k, \ \psi_k(y_2)=m_k, \ \psi_k(y_3)=\alpha m_k.
\label{HIPPY}
\end{equation}

\subsubsection{Subsets of integers and some properties}

For our construction, it is useful to introduce a suitable countable set of matrices $M_n$. They define a reordering of the nonnegative integers and their union contains all of them without repetitions. One has

\begin{equation}
M_n := 
\left[\begin{array}{ccccc}
y_n & y_n+r_n & y_n+2r_n & \dots & y_n+(c_n-1)r_n \\
y_n+1 & y_n+r_n+1 & y_n+2r_n+1 & \dots & y_n+(c_n-1)r_n+1 \\
y_n+2 & y_n+r_n+2 & y_n+2r_n+2 & \dots & y_n+(c_n-1)r_n+2 \\
\dots & \dots & \dots & \dots & \dots \\
y_n+r_n-1 & y_n+2r_n-1 & y_n+3r_n-1 & \dots & y_n+c_nr_n-1
\end{array}\right]
\label{URK}
\end{equation}
where 
\begin{itemize}
\item $r_n$ is the number of rows. It also represents the distance between two successive numbers in the same row; 
\item $c_n$ is the number of columns (and note that inside a column the distance between successive numbers is always equal to $1$);
\item the first matrix $M_1$ begins with the first non-negative number, i.e. $y_1=0$;
\item the scalar $y_n$ is the element in position $(1,1)$ of the matrix $M_n$. It also easily holds that $y_{n+1}=y_n+c_nr_n$;
\item the matrix $M_n$ thus contains all the integers between $y_n$ and $y_{n+1}-1$.
\end{itemize}
Therefore, we can write
$$
y_1=0, \ y_n=c_1r_1+c_2r_2+\dots+c_{n-1}r_{n-1}, \ n \ge 2.
$$
In what follow, for reasons which will be clear later on, we choose 
$$
r_n=135 \ 2^{n-1}, \ c_n=2^{n-1}
$$
so that
$$
y_n=135(1+4+4^2+\dots+4^{n-2})=45(4^{n-1}-1) \simeq 45 \ 4^{n-1}.
$$

\noindent It is now needed to analyze the distance between the square roots of the half of two successive numbers in the same $(h+1)^{th}$ row in $M_n$. By using $\sqrt{1+x}\simeq 1+\frac{x}{2}$ if $|x|<<1$, for $k=1,2,\dots,c_n-1$ and $h=0,1,,\dots,r_n-1$ one has
{\footnotesize \begin{eqnarray*}\
\frac{1}{\sqrt{2}}[\sqrt{y_n+h+kr_n}-\sqrt{y_n+h+(k-1)r_n}]&=&\frac{1}{\sqrt{2}}\left\{\sqrt{y_n+h+kr_n}\left[1-\sqrt{\frac{y_n+h+(k-1)r_n}{y_n+h+kr_n}}\right]\right\}\\
&=&\frac{1}{\sqrt{2}}\left\{\sqrt{y_n+h+kr_n}\left[1-\sqrt{1-\frac{r_n}{y_n+h+kr_n}}\right]\right\}\\
&\simeq&\frac{r_n}{2\sqrt{2}\sqrt{y_n+h+kr_n}}\\
&\ge&\frac{r_n}{2\sqrt{2y_{n+1}}}\\
&=&\frac{135}{4\sqrt{90}} \simeq 3.56, \ \mbox{for} \ n\rightarrow+\infty.
\end{eqnarray*}
}

\noindent So, as $n$ grows to infinity, for the 
points $P_{k-1}$ and $P_k$ present in the same row of $M_n$ one has
\begin{equation}
\sqrt{\frac{P_{k}}{2}}-\sqrt{\frac{P_{k-1}}{2}} \gtrsim 3.56.
\label{UFG}
\end{equation}

\noindent The above equation is important because $\sqrt{\frac{k}{2}}$ are the maximum points for the main functions $\psi_k(x)$ in \eqref{Gauss}. Accordingly to (\ref{HIPPY}) and (\ref{UFG}), it holds that
$$
\alpha \lesssim  10^{-11(k_1-k_2)^2} 
$$
for the relative influence of the $k_1$th peak on the $k_2$th one and conversely.
Thus the contribution to the following algebraic sum
\begin{equation}
g(x):=\frac{1}{\sqrt[4]{2\pi P_1}}\ e^{-2\left(x-\sqrt{\frac{P_1}{2}}\right)^2}\pm\frac{1}{\sqrt[4]{2\pi P_2}}\ e^{-2\left(x-\sqrt{\frac{P_2}{2}}\right)^2}\pm\dots\pm\frac{1}{\sqrt[4]{2\pi P_{c_n}}}\ e^{-2\left(x-\sqrt{\frac{P_{c_n}}{2}}\right)^2}
\label{SIS}
\end{equation}
of all the terms but the first one are negligible in a neighborhood of $\sqrt{\frac{P_1}{2}}$. Both the maximum point and the maximum value are almost unaffected by all the terms after the first one, regardless of the sequence of chosen signs $\pm$. This is equivalent to saying that
\begin{equation}
\|g\|_{\infty}\simeq\frac{1}{\sqrt[4]{2\pi P_1}} \ \Rightarrow \ \|g\|_{\infty}^2\simeq\frac{1}{\sqrt{2\pi P_1}}.
\label{FGH}
\end{equation}


\begin{remark} 
Now we can see the reason for the previous choice of $r_n$. The number $135$ is strongly associated with making $\alpha$ negligible. 
Furthermore, the first local maximum is larger than the others (in absolute value), so the absolute maximum should be looked for near it. However, as $n$ tends to infinity, the first peaks tend to become more similar, and the absolute maximum may fall at a different point. 
In fact, the contribution to the first peak of all the terms but the second can be neglected as they are of the order of $10^{-44},10^{-99},\ldots$. Hence, only the contribution of the second, of order $10^{-11}$, needs to be taken into account, while there is a double contribution of $2 \times 10^{-11}$ to the second peak, due to the first and third terms. It follows that, when the (relative) difference of the first two peaks becomes less than $10^{-11}$, the maximum value is obtained around the second peak in the case of all positive signs (similar considerations apply when dealing with different signs). When the (relative) difference becomes less than $10^{-44}$, such small corrections begin to play a role and the maximum can move to the third peak, and so on. But all these corrections, which are always relative and not absolute, remain negligible and do not affect the value of the norm. These arguments make it easy to prove that the sup-norm is not affected in practice, even if the real maximum does not coincide with any of the peaks of the summed functions.  
\end{remark} 

\subsubsection{First step in building the new functions}

Our aim is now to replace the basis functions $\psi_k(x)=\sqrt{\frac{2^k}{k!}}x^ke^{-x^2}$, forming the starting expansion \eqref{BasicExp}, with a new linearly independent set $\phi_k(x)$
with the following properties
\begin{itemize}
\item they give rise to the following kernel expansion
$$
K(x,y)=\sum_{k=0}^{+\infty} \ \psi_i(x)\psi_i(y)=\sum_{k=0}^{+\infty} \ \phi_i(x)\phi_i(y);
$$
\item their $\infty-$norms are such that
$$
\sum_{k=0}^{n} \ \|\phi_i\|_{\infty}^2 \ \mbox{diverges as} \ log(n)
$$
and
$$
\sum_{k=0}^{+\infty} \ \|\phi_i\|_{\infty}^{2p}<+\infty, \ \forall p>1.
$$
So, rewriting $K(x,y)$ in the following normalized way
\begin{eqnarray*}
K(x,y)&=&\sum_{k=0}^{+\infty} \ \mu_i \ \frac{\psi_i(x)}{\|\psi_i\|_{\infty}}\ \frac{\psi_i(y)}{\|\psi_i\|_{\infty}}\\
&=& \sum_{k=0}^{+\infty} \ \lambda_i \ \frac{\phi_i(x)}{\|\phi_i\|_{\infty}}\ \frac{\phi_i(y)}{\|\phi_i\|_{\infty}}
\end{eqnarray*}
the best possible improvement is achieved: the $\mu_i$ fall in $\ell_p$ for all $p>2$ while the $\lambda_i$ belong to $\ell_p$ for all $p>1$. Furthermore,  the sum of the $\lambda_i$, i.e. their norm in $\ell_1$, diverges very slowly. This is the best possible scenario since we have proved
that the constraint on the basis functions prevents summability of the expansion coefficients.
\end{itemize}
First, we replace the functions defined by the indexes in any single row of the matrix $M_n$. For simplicity sake, the original functions are denoted by by $A_1,\dots, A_{c_n}$. Next, defining
$$
B_1(x):=\frac{1}{\sqrt{2}}[A_1(x)+A_2(x)], \ B_2(x):=\frac{1}{\sqrt{2}}[A_1(x)-A_2(x)]
$$
it follows that
\begin{eqnarray*}
A_1(x)A_1(y)+A_2(x)A_2(y)&=&B_1(x)B_1(y)+B_2(x)B_2(y)\\
\text{span}[A_1(x), A_2(x)]&=&\text{span}[B_1(x), B_2(x)].
\end{eqnarray*}
So, $B_1,B_2$ remain linearly independent if $A_1,A_2$ are. We then apply the following algorithm to the $2^{n-1}$ functions $A_1,\dots,A_{c_n}$: 
\begin{itemize}
\item set $A_1^{(1)}:=A_1,\dots,A_{c_n}^{(1)}=A_{c_n}$;
\item starting from $A_1^{(h)},\dots,A_{c_n}^{(h)}$, for
 $h=1,2,\dots,n-1$ build recursively 
$$
\begin{array}{lcclccl}
A_1^{(h+1)}&:=&\frac{1}{\sqrt{2}} \ (A_1^{(h)}+A_2^{(h)}), &A_{2^{n-2}+1}^{(h+1)}&:=&\frac{1}{\sqrt{2}} \ (A_1^{(h)}-A_2^{(h)}),\\
A_2^{(h+1)}&:=&\frac{1}{\sqrt{2}} \ (A_3^{(h)}+A_4^{(h)}), &A_{2^{n-2}+2}^{(h+1)}&:=&\frac{1}{\sqrt{2}} \ (A_3^{(h)}-A_4^{(h)}),\\
&\dots&\\
A_{2^{n-2}}^{(h+1)}&:=&\frac{1}{\sqrt{2}} \ (A_{2^{n-1}-1}^{(h)}+A_{2^{n-1}})^{(h)}, &A_{2^{n-1}}^{(h+1)}&:=&\frac{1}{\sqrt{2}} \ (A_{2^{n-1}-1}^{(h)}-A_{2^{n-1}}^{(h)}).
\end{array}
$$
We still have $c_n$ linearly independent functions, keeping the following sum invariant
$$
A_1^{(h)}(x)A_1^{(h)}(y)+\dots+A_{c_n}^{(h)}(x)A_{c_n}^{(h)}(y)=A_1^{(h+1)}(x)A_1^{(h+1)}(y)+\dots+A_{c_n}^{(h+1)}(x)A_{c_n}^{h+1)}(y)
$$
with the $A_i^{(h+1)}$ sums of some $A_i^{(h)}$ with different signs (coefficients $\pm 1$), apart from the multiplicative diminishing factor $\frac{1}{\sqrt{2}}$. Moreover
$$
\begin{array}{lcl}
\text{span}[A_1^{(h+1)},A_{2^{n-2}+1}^{(h+1)}]&=&\text{span}[A_1^{(h)},A_2^{(h)}]\\
\text{span}[A_2^{(h+1)},A_{2^{n-2}+2}^{(h+1)}]&=&\text{span}[A_3^{(h)},A_4^{(h)}]\\
&\dots&\\
\text{span}[A_{2^{n-2}}^{(h+1)},A_{2^{n-1}}^{(h+1)}]&=&\text{span}[A_{2^{n-1}-1}^{(h)},A_{2^{n-1}}^{(h)}].
\end{array}
$$
\item without losing the linear independence property the procedure can be repeated only $(n-1)$ times. One then obtains something like
$$
A_1^{(n)}(x)A_1^{(n)}(y)+\dots+A_{c_n}^{(n)}(x)A_{c_n}^{(n)}(y)=A_1(x)A_1(y)+\dots+A_{c_n}(x)A_{c_n}(y)
$$
with the $A_i^{(n)}$ linearly independent and
$$
\text{span}[A_1^{(n)}(x),\dots,A_{c_n}^{(n)}(x)]=\text{span}[A_1(x),\dots,A_{c_n}(x)].
$$
\end{itemize}
Finally note that each $A_i^{(n)}$ is a linear combination of all the $A_i$ with different signs, apart for the diminishing multiplicative factor $\frac{1}{(\sqrt{2})^{n-1}}=\frac{1}{\sqrt{c_n}}$ (each step leads to a decreasing factor of $\frac{1}{\sqrt{2}}$). Recalling (\ref{FGH}), this means that the $\infty-$norms of the $A_i^{(n)}$ are made smaller than that of $A_1$ by the multiplicative factor $\frac{1}{\sqrt{c_n}}$, so that
\begin{equation}
A_i^{(n)}(x)=\frac{1}{\sqrt{c_n}}[A_1(x)\pm A_2(x) \pm \dots A_{c_n}(x)] \ \Rightarrow \ \|A_i^{(n)}\|_{\infty}^2 \simeq \frac{1}{c_n} \ \|A_1\|_{\infty}^2.
\label{EQQE}
\end{equation}
This last equation (\ref{EQQE}) will be the key to the solution of our problem.

\subsubsection{An illustrative example}

To better understand the procedure reported in the previous section,
the construction process is illustrated for $n=4$:
\begin{itemize}
\item set $A_i^{(1)}:=A_i$ for $i=1,2,\dots,8$;
\item define 
{\footnotesize \begin{eqnarray*}
A_1^{(2)}&=&\frac{1}{\sqrt{2}}(A_1+A_2), \ A_2^{(2)}=\frac{1}{\sqrt{2}}(A_3+A_4), \ A_3^{(2)}=\frac{1}{\sqrt{2}}(A_5+A_6), \ A_4^{(2)}=\frac{1}{\sqrt{2}}(A_7+A_8),\\
A_5^{(2)}&=&\frac{1}{\sqrt{2}}(A_1-A_2), \ A_6^{(2)}=\frac{1}{\sqrt{2}}(A_3-A_4), \ A_7^{(2)}=\frac{1}{\sqrt{2}}(A_5-A_6), \ A_8^{(2)}=\frac{1}{\sqrt{2}}(A_7-A_8)
\end{eqnarray*}}
and note that
{\begin{eqnarray*}
\text{span}[A_1^{(2)},A_5^{(2)}]&=&\text{span}[A_1,A_2], \ \text{span}[A_2^{(2)},A_6^{(2)}]=\text{span}[A_3,A_4],\\
\text{span}[A_3^{(2)},A_7^{(2)}]&=&\text{span}[A_5,A_6], \ \text{span}[A_4^{(2)},A_8^{(2)}]=\text{span}[A_7,A_8]
\end{eqnarray*}}
\item now build 
{\footnotesize \begin{eqnarray*}
A_1^{(3)}&=&\frac{1}{\sqrt{2}}(A_1^{(2)}+A_2^{(2)})=\frac{1}{2}(A_1+A_2+A_3+A_4), \ A_2^{(3)}=\frac{1}{\sqrt{2}}(A_3^{(2)}+A_4^{(2)})=\frac{1}{2}(A_5+A_6+A_7+A_8),\\
A_3^{(3)}&=&\frac{1}{\sqrt{2}}(A_5^{(2)}+A_6^{(2)})=\frac{1}{2}(A_1-A_2+A_3-A_4), \ A_4^{(3)}=\frac{1}{\sqrt{2}}(A_7^{(2)}+A_8^{(2)})=\frac{1}{2}(A_5-A_6+A_7-A_8),\\
A_5^{(3)}&=&\frac{1}{\sqrt{2}}(A_1^{(2)}-A_2^{(2)})=\frac{1}{2}(A_1+A_2-A_3-A_4), \ A_6^{(3)}=\frac{1}{\sqrt{2}}(A_3^{(2)}-A_4^{(2)})=\frac{1}{2}(A_5+A_6-A_7-A_8),\\
A_7^{(3)}&=&\frac{1}{\sqrt{2}}(A_5^{(2)}-A_6^{(2)})=\frac{1}{2}(A_1-A_2-A_3+A_4), \ A_8^{(3)}=\frac{1}{\sqrt{2}}(A_7^{(2)}-A_8^{(2)})=\frac{1}{2}(A_5-A_6-A_7+A_8)
\end{eqnarray*}}
and one has
{\small \begin{eqnarray*} 
\text{span}[A_1^{(3)},A_3^{(3)},A_5^{(3)},A_7^{(3)}]&=&\text{span}[A_1,A_2,A_3,A_4],  \\
\text{span}[A_2^{(3)},A_4^{(3)},A_6^{(3)},A_8^{(3)}]&=&\text{span}[A_5,A_6,A_7,A_8]
\end{eqnarray*}}
\item finally, let 
\begin{eqnarray*}
A_1^{(4)}&=&\frac{1}{\sqrt{2}}(A_1^{(3)}+A_2^{(3)})=\frac{1}{2\sqrt{2}}(A_1+A_2+A_3+A_4+A_5+A_6+A_7+A_8),\\
A_2^{(4)}&=&\frac{1}{\sqrt{2}}(A_3^{(3)}+A_4^{(3)})=\frac{1}{2\sqrt{2}}(A_1-A_2+A_3-A_4+A_5-A_6+A_7-A_8),\\
A_3^{(4)}&=&\frac{1}{\sqrt{2}}(A_5^{(3)}+A_6^{(3)})=\frac{1}{2\sqrt{2}}(A_1+A_2-A_3-A_4+A_5+A_6-A_7-A_8),\\
A_4^{(4)}&=&\frac{1}{\sqrt{2}}(A_7^{(3)}+A_8^{(3)})=\frac{1}{2\sqrt{2}}(A_1-A_2-A_3+A_4+A_5-A_6-A_7+A_8),\\
A_5^{(4)}&=&\frac{1}{\sqrt{2}}(A_1^{(3)}-A_2^{(3)})=\frac{1}{2\sqrt{2}}(A_1+A_2+A_3+A_4-A_5-A_6-A_7-A_8),\\
A_6^{(4)}&=&\frac{1}{\sqrt{2}}(A_3^{(3)}-A_4^{(3)})=\frac{1}{2\sqrt{2}}(A_1-A_2+A_3-A_4-A_5+A_6-A_7+A_8),\\
A_7^{(4)}&=&\frac{1}{\sqrt{2}}(A_5^{(3)}-A_6^{(3)})=\frac{1}{2\sqrt{2}}(A_1+A_2-A_3-A_4-A_5-A_6+A_7+A_8),\\
A_8^{(4)}&=&\frac{1}{\sqrt{2}}(A_7^{(3)}-A_8^{(3)})=\frac{1}{2\sqrt{2}}(A_1-A_2-A_3+A_4-A_5+A_6+A_7-A_8)
\end{eqnarray*}
\noindent with
$$
\text{span}[A_1^{(4)},A_2^{(4)},\dots,A_8^{(4)}]=\text{span}[A_1,A_2,\dots,A_8].
$$
\end{itemize}
The signs of $A_1,\dots,A_8$ in the various $A_1^{(4)},\dots,A_8^{(4)}$ are chosen accordingly to the following table
\begin{equation}\label{Table1}
\left[\begin{array}{ccccccccc}& A_1 & A_2 & A_3 & A_4 & A_5 & A_6 & A_7 & A_8 \cr A_1^{(4)} &+ & + & + & + & + & + & + & +\cr A_2^{(4)} & + & - & + & - & + & - & + & -\cr A_3^{(4)} & + & + & - & - & + & + & - & -\cr A_4^{(4)} & + & - & - & + & + & - & - & +\cr A_5^{(4)} & + & + & + & + & - & - & - & -\cr A_6^{(4)} & + & - & + & - & - & + & - & +\cr A_7^{(4)} & + & + & - & - & - & - & + & +\cr A_8^{(4)} & + & - & - & + & - & + & + & -\end{array}\right]
\end{equation}

\noindent Applying this procedure to the first row of $M_4$ results in the following functions:

\begin{eqnarray*}
A_1(x)&=&\psi_{y_4}(x)=\psi_{2835}(x)=\sqrt{\frac{2^{2835}}{2835!}}x^{2835}e^{-x^2}\\
A_2(x)&=&\psi_{y_4+r_4}(x)=\psi_{3915}(x)=\sqrt{\frac{2^{3915}}{3915!}}x^{3915}e^{-x^2}\\
A_3(x)&=&\psi_{y_4+2r_4}(x)=\psi_{4995}(x)=\sqrt{\frac{2^{4995}}{4995!}}x^{4995}e^{-x^2}\\
A_4(x)&=&\psi_{y_4+3r_4}(x)=\psi_{6075}(x)=\sqrt{\frac{2^{6075}}{6075!}}x^{6075}e^{-x^2}\\
A_5(x)&=&\psi_{y_4+4r_4}(x)=\psi_{7155}(x)=\sqrt{\frac{2^{7155}}{7155!}}x^{7155}e^{-x^2}\\
A_6(x)&=&\psi_{y_4+5r_4}(x)=\psi_{8235}(x)=\sqrt{\frac{2^{8235}}{8235!}}x^{8235}e^{-x^2}\\
A_7(x)&=&\psi_{y_4+6r_4}(x)=\psi_{9315}(x)=\sqrt{\frac{2^{9315}}{9315!}}x^{9315}e^{-x^2}\\
A_8(x)&=&\psi_{y_4+7r_4}(x)=\psi_{10395}(x)=\sqrt{\frac{2^{10395}}{10395!}}x^{10395}e^{-x^2}.
\end{eqnarray*}



\subsection{Properties of the new functions}

According to what seen in the previous section, we replace the original $\psi_k(x)$ corresponding to the indexes of any single row in $M_n$ with new $\phi_k(x)$ (corresponding to the same indexes) in such a way that linear independence still holds. 
In doing this, we deal with finite subsets of ${\mathbb N}$ satisfying
$$
{\mathcal I}_k \cap {\mathcal I}_h=\emptyset \ \mbox{if} \ k \ne h, \ \ \ \cup_{k=1}^{+\infty} \ {\mathcal I}_k = {\mathbb N}
$$
and we make a basis change within each of the corresponding functions set 
to preserve independence. In view of \eqref{EQQE}, for any single row in $M_n$ it holds that
$$
\lambda(n,h,k):=\|\phi_{y_n+h+kr_n}\|_{\infty}^2\simeq  \frac{1}{c_n}, \  \|\psi_{y_n+h}\|_{\infty}^2
$$
for $h=0,1,2,\dots,r_n-1$ and $k=0,1,2,\dots,c_n-1$.
Recalling that
$$
\|\psi_{y_n+h}\|_{\infty}^2\simeq\frac{1}{\sqrt{2\pi (y_n+h)}}\le\frac{1}{\sqrt{2\pi y_n}},
$$
by summing up the $p-$power ($p\ge 1$) of the $\lambda(n,h,k)$ corresponding to the same row, one obtains
\begin{equation}
\sum_{k=0}^{c_n-1} \ \lambda^p(n,h,k)\simeq c_n \times \frac{\|\psi_{y_n+h}\|_{\infty}^{2p}}{c_n^p}\le\frac{1}{(\sqrt{2\pi y_n})^p}c_n^{1-p}.
\label{UFFAH}
\end{equation}
Furthermore, by summing again the contribution of each row, one has
\begin{eqnarray*}
G(n,p)&:=&\sum_{h=0}^{r_n-1} \ \sum_{k=0}^{c_n-1} \ \lambda^p(n,h,k)\simeq \frac{r_n}{(\sqrt{2\pi y_n})^p}c_n^{1-p}\\
&=&\frac{135 \ 4^{p-1}}{(\sqrt{90\pi})^p} \ \frac{1}{(4^{p-1})^n}\\
&:=& \frac{A( p )}{b^n( p )}
\end{eqnarray*}
which represents the whole contribution of the matrix $M_n$ for $n$ large enough
(and we are only interested in the asymptotic behavior). The cases $p=1$ and $p>1$ lead, respectively, to
$$
p=1 \ \Rightarrow \ G(n,1)\simeq \frac{135}{\sqrt{90\pi}} \simeq 8 \ \ \ \ (A( 1 )\simeq\frac{135}{\sqrt{90\pi}}, \ b( 1 )=1),
$$
and
$$
p>1 \ \Rightarrow \ G(n,p)=\frac{A( p )}{b^n( p )}, \ A( p )>0, \ b( p )>1.
$$
This implies that
$$
\sum_{k=0}^{+\infty} \ \|\phi_k\|^2=\sum_{n=1}^{+\infty} \ G(n,1)=+\infty
$$
and
$$
\sum_{k=0}^{+\infty} \ \|\phi_k\|^{2p}=\sum_{n=1}^{+\infty} \ G(n,p)=\frac{A( p )b( p )}{b( p )-1}<+\infty, \ p>1.
$$
This shows that the new weights do not belong to $\ell_1$ but fall in $\ell_p$ for any $p>1$.
\medskip

\noindent As far as the divergence speed for $p=1$ and the convergence rate for $p>1$ is concerned, let us consider the situation $\lambda_k^*=\frac{D}{k}$. In this case the contribution of $M_n$ would be
\begin{eqnarray*}
\sum_{h=y_n}^{y_{n+1}-1} \ \lambda_h^* &=& \sum_{h=y_n}^{y_{n+1}-1} \ \frac{D}{k} \\
&\simeq& \int_{y_n}^{y_{n+1}} \ \frac{D}{x}dx=D[\log[y_{n+1}]-\log[y_n]]\\
&\simeq& D \ \log[\frac{45 \ 4^n}{45 \ 4^{n-1}}]\\
&=& D \ \log(4) \simeq 8
\end{eqnarray*}
for a suitable $D\simeq\frac{135}{\sqrt{90\pi} \ log(4)}$. This means that, from an average viewpoint, the actual $\lambda_k$ behave as they were equal to $\lambda_k^*=\frac{D}{k}$. Therefore, from
$$
\sum_{k=0}^n \ \lambda_k \simeq \sum_{k=1}^n \ \frac{D}{k} \simeq \int_1^n \ \frac{D}{x}dx=D \ log(n)
$$
one can see that the sum increases very slowly (logarithmically). This is only true on average, as the weighting sequence actually attempts to track the values $\frac{D}{k}$ by fluctuating around them. This is confirmed even for the $\ell_p$ case where, in fact, one has
\begin{eqnarray*}
 \sum_{h=y_n}^{y_{n+1}-1} \ (\lambda_h^*)^p&=&\sum_{h=y_n}^{y_{n+1}-1} \ \frac{D^p}{k^p} \simeq \int_{y_n}^{y_{n+1}} \ \frac{D^p}{x^p}dx\\
 &=&\frac{D^p}{1-p}x^{1-p}|_{y_n}^{y_{n+1}}\\
 &=&\frac{D^p (4^{p-1}-1)}{(p-1)45^{p-1}}\frac{1}{(4^{p-1})^n}\\
 &=&\frac{A( p )}{b^n( p )}.
\end{eqnarray*}

\noindent Two final comments are in order. The fact that we have found the sequence $\frac{A( p )}{b^n( p )}$ which exponentially converges to zero could be misleading. The convergence rate is not exponential and similar to that of the sequence $(\lambda_k^*)^p=\frac{1}{k^p}$, and it holds something like
$$
\sum_{k=0}^n \ \lambda_k^p \simeq \sum_{k=1}^n \ \frac{D^p}{k^p}\simeq \int_1^n \ \frac{D^p}{x^p}dx=D^p\left(1-\frac{1}{p-1}\ \frac{1}{n^{p-1}}\right).
$$
The reason why we dealt with an exponential rate is because we have considered the total contribution of the matrix $M_n$.\\
Finally, we can see from (\ref{UFFAH}) with $p=1$ that the effect of introducing the new functions is to eliminate most of the original $m_k^2$. In fact, only the $m_k^2$ corresponding to the first column of each matrix $M_n$ need to be taken into account, since the sum of the new weights corresponding to the whole $M_n$ is almost equal to the sum of the old ones corresponding only to the first columns. This means that a large part of the old weights has been discarded from the first list, and this is why we were able to improve the $\ell_p-$property from $p>2$ to $p>1$.

\vskip 0.2in


\end{document}